\documentclass[sigconf,natbib=true]{acmart}
\AtBeginDocument{%
  }

\setcopyright{acmlicensed}
\copyrightyear{2025}
\acmYear{2025}
\acmDOI{XXXXXXX.XXXXXXX}
\acmConference[CIKM '25]{the 34th ACM International Conference on Information and Knowledge Management (CIKM ’25), November 10--14, 2025, Seoul, Republic of Korea}{November 10--04, 2025}{Seoul, Republic of Korea.}
\acmISBN{XXX-X-XXXX-XXXX-X}

\begin{document}

\title{Let Multimodal Embedders Learn When to Augment Query via Adaptive Query Augmentation}


\author{Wongyu Kim}
\authornote{Corresponding author}
\email{wgkim@ncsoft.com}
\affiliation{%
  \institution{NC AI}
  \city{Seongnam}
  \country{Republic of Korea}
}

\author{Hochang Lee}
\email{hochang@ncsoft.com}
\affiliation{%
  \institution{NC AI}
  \city{Seongnam}
  \country{Republic of Korea}
}

\author{Sanghak Lee}
\email{sanghaklee@ncsoft.com}
\affiliation{%
  \institution{NC AI}
  \city{Seongnam}
  \country{Republic of Korea}
}

\author{Yoonsung Kim}
\email{yoonsungkim@ncsoft.com}
\affiliation{%
  \institution{NC AI}
  \city{Seongnam}
  \country{Republic of Korea}
}

\author{Jaehyun Park}
\email{jaehyunpark@ncsoft.com}
\affiliation{%
  \institution{NC AI}
  \city{Seongnam}
  \country{Republic of Korea}
}

\renewcommand{\shortauthors}{Kim et al.}

\begin{abstract}
Query augmentation makes queries more meaningful by appending further information to the queries to find relevant documents. Current studies have proposed Large Language Model (LLM)-based embedders, which learn representation for embedding and generation for query augmentation in a multi-task manner by leveraging the generative capabilities of LLM. During inference, these jointly trained embedders have conducted query augmentation followed by embedding, showing effective results. However, augmenting every query leads to substantial embedding latency and query augmentation can be detrimental to performance for some queries. Also, previous methods have not been explored in multimodal environments. To tackle these problems, we propose M-Solomon, a universal multimodal embedder that can adaptively determine when to augment queries. Our approach first divides the queries of the training datasets into two groups at the dataset level. One includes queries that require augmentation and the other includes queries that do not. Then, we introduces a synthesis process that generates appropriate augmentations for queries that require them by leveraging a powerful Multimodal LLM (MLLM). Next, we present adaptive query augmentation. Through this step, M-Solomon can conduct query augmentation only when necessary by learning to generate synthetic augmentations with the prefix \textit{/augment} for queries that demand them and to generate the simple string \textit{/embed} for others. Experimental results showed that M-Solomon not only surpassed the baseline without augmentation by a large margin but also outperformed the baseline that always used augmentation, providing much faster embedding latency. 
\end{abstract}



\maketitle

\begin{figure}[h]
  \centering
  \includegraphics[width=\linewidth]{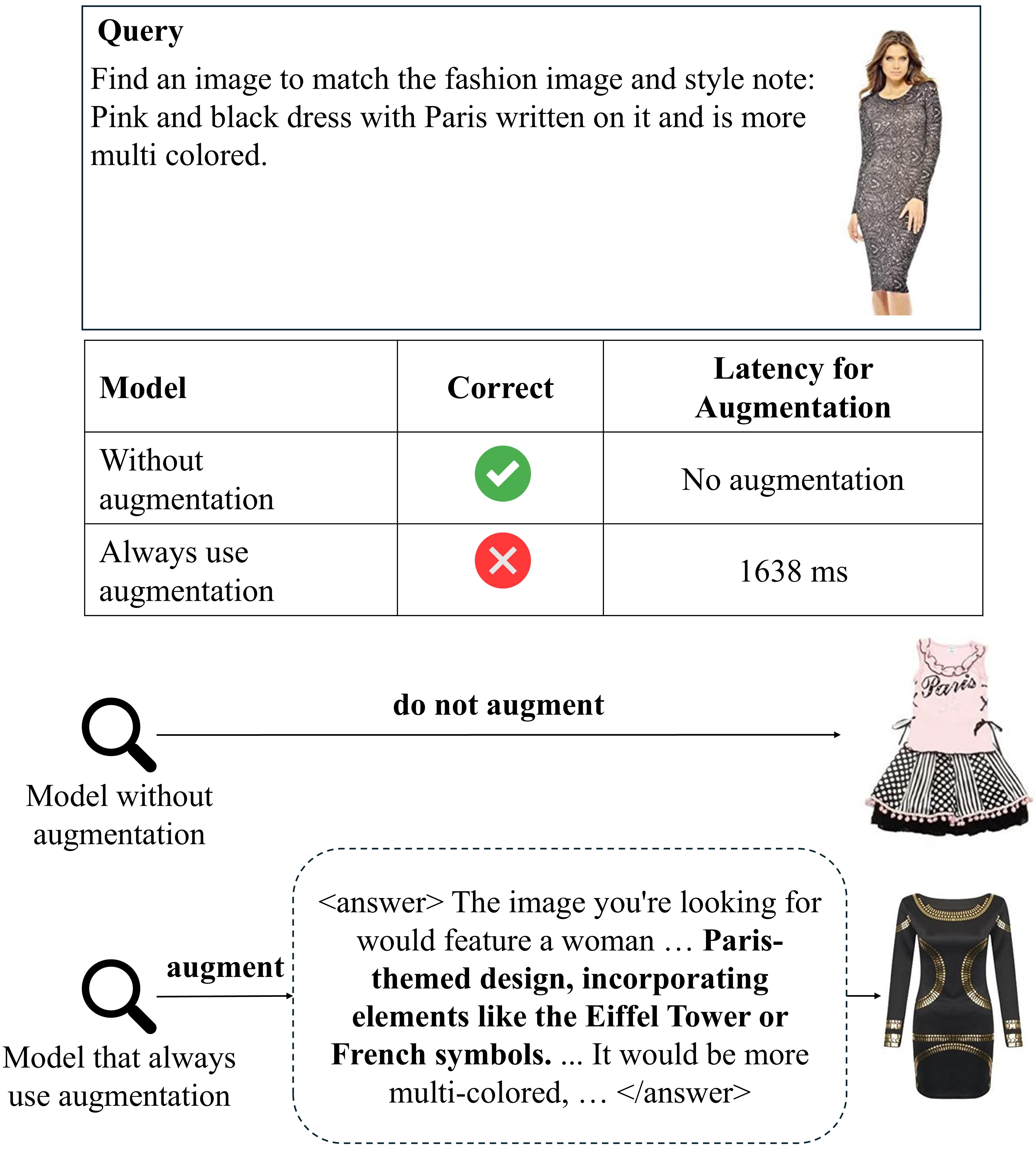}
  \caption{An example of FashionIQ dataset in MMEB benchmark \cite{jiang2024vlm2vec} for the pilot study.}
  \Description{A woman and a girl in white dresses sit in an open car.}
  \label{fig:pilot}
\end{figure}

\section{Introduction}
In many embedding tasks, query augmentation has been used for retrieving relevant documents by appending useful information to queries. Recent studies have proposed Large Language Model (LLM)-based embedders which jointly learn representation for embedding and generation for query augmentation by leveraging the generative capabilities of LLM \cite{yan2025o1, tang2025think}. 
During inference, these embedders trained in the multi-task manner have conducted query augmentation followed by embedding, which has led to more meaningful query representations and stronger performance. 
However, previous studies did not consider the following three points: (1) Augmenting every query leads to significant embedding latency. (2) Query augmentation can degrade performance for some queries. (3) The effectiveness has not been demonstrated in multimodal environments. To verify these challenges, we conducted a pilot study as shown in Figure~\ref{fig:pilot}. We trained two models: one that performs only embedding, similar to typical embedders, and the other that carries out query augmentation before embedding. The former employed widely used contrastive loss for training \cite{chen2020simple}, while the latter is trained by referring to \cite{yan2025o1}. In Figure~\ref{fig:pilot}, the model without augmentation quickly retrieved the relevant image document, while the model that always uses augmentation did not. The bolded parts of the augmentation were generated by misinterpreting and exaggerating the query. They pointed to finding clothes with Paris-related designs like the Eiffel Tower instead of just the word `Paris', which was expected to hinder finding the positive document.

To tackle these challenges, based on the results of the pilot study, we propose M-Solomon, a universal multimodal embedder that can adaptively determine when to augment queries. 
Referring to \cite{jiang2025think}, our approach first divides the queries of the training datasets into two groups at the dataset level: one that contains queries that require augmentation and the other that contains queries that do not require augmentation. 
Subsequently, by leveraging a high-performance Multimodal LLM (MLLM), we introduce a synthesis process that generates appropriate answers for queries that require augmentation and these answers are regarded as augmentations \cite{yan2025o1}. Lastly, inspired by adaptive generation between thinking and non-thinking modes \cite{zhang2025adaptthink, fang2025thinkless, lou2025adacot}, we present jointly learning adaptive query augmentation in addition to learning representation for embedding. M-Solomon can augment queries only when necessary by learning to generate synthetic augmentations with the prefix \textit{/augment} for queries that demand augmentation and to generate the simple string \textit{/embed} for others. Producing these tokens at the beginning makes M-Solomon decide whether to augment each query.

Experimental results on MMEB benchmark \cite{jiang2024vlm2vec} showed that M-Solomon substantially outperformed the baseline without augmentation by adaptively augmenting queries. Also, M-Solomon exhibited better performance compared to the baseline that constantly used augmentation, achieving significantly faster embedding latency.

\section{Related Work}
\textbf{Joint Training of Embedding and Query Augmentation.} Recent studies \cite{yan2025o1, tang2025think} have developed embedders to learn embedding and query augmentation in a multi-task manner. By augmenting queries before embedding, query representations have become more informative. 
However, when augmenting every query, embedding latency increases significantly and performance can decline. Also, previous studies have not considered multimodal environments. Our approach addresses these problems.

\textbf{Multimodal Embedders.} Since the release of MMEB \cite{jiang2024vlm2vec}, a multimodal embedding training collection and benchmark, numerous multimodal embedders have been developed by using various techniques such as efficient GPU utilization to increase batch size \cite{jiang2024vlm2vec}, data synthesis \cite{zhou2024megapairs, chen2025mme5, liu2025idmr}, hard negative sampling \cite{schneider2025abc, gu2025breaking}, contrastive-autoregressive finetuning \cite{yu2025cafe}, distillation \cite{gu2025breaking, thirukovalluru2025breaking}, prompt refinement \cite{thirukovalluru2025breaking, kong2025modality}, modality completion module \cite{qin2025unimoco}, and optimization of contrastive loss \cite{lan2025llave, thirukovalluru2025breaking, kong2025modality, xue2025improve}. We propose a novel method to improve performance by adaptively augmenting queries.

\textbf{Adaptive Generation.} Recent LLMs have demonstrated that, for certain questions, generating direct answers with non-thinking mode can be more effective and efficient than answering with thinking mode \cite{ma2025reasoning, yang2025qwen3}. Several methods have been proposed to adaptively generate responses by automatically selecting the appropriate strategy between thinking and non-thinking modes for each question, enabling more effective and efficient test-time scaling \cite{zhang2025adaptthink, fang2025thinkless, lou2025adacot, jiang2025think, yue2025hybrid, wang2025pats}. Inspired by these adaptive generation methods, we introduce adaptive query augmentation.

\section{Methodology}

\textbf{Task Definition.} The training dataset collection contains $[D^1_A, D^2_A,$ $\dots, D^a_A, D^1_E, D^2_E, \dots, D^e_E]$, where $D^u_A$ and $D^v_E$ respectively denote a dataset that contains queries requiring augmentation and a dataset that contains queries not requiring augmentation, $u$ and $v$ are the indices for $D_A$ and $D_E$, and $a$ and $e$ each mean the sizes of $D_A$ and $D_E$. Each dataset has $(q^i, g^i, p^i, n^i_1, n^i_2, ..., n^i_m)$ samples, where $q^i$, $p^i, n^i_k$ and $m$ each indicate a query, a positive document, a hard negative document and the nubmer of $n$, and $i$ and $k$ are the indices for samples and hard negative documents. $g^i$ denotes a synthetic augmentation with the prefix \textit{/augment} if it belongs to $D^u_A$ or the the simple string \textit{/embed} if it belongs to $D^v_E$. The modalities of $q^i, p^i, n^i_k$ are text, image, or interleaved text and image, while the modality of $g^i$ is text. M-Solomon aims to learn not only representing the augmented query ($q$ with $g$) and retrieving the positive document ($p$) but also adaptively generating the augmentation ($g$). During evaluation, on the benchmark that includes datasets $[D^1_B, D^2_B, ..., D^b_B]$, M-Solomon aims to find the positive document among document candidates by using the augmented query. $b$ means the number of $D_B$.

\subsection{Query Augmentation Synthesis} \label{subsec:synthesis} The core capability of M-Solomon lies in adaptively determining when to augment queries. Therefore, identifying which queries benefit from augmentation and synthesizing augmentations for those queries are necessary to construct the dataset collection that makes M-Solomon acquire that capability. Referring to \cite{jiang2025think}, we first divide the queries of the training datasets at the dataset level by utilizing the models in the pilot study (Figure~\ref{fig:pilot})\footnote{In the pilot study, the model without augmentation is trained by using contrastive loss in Equation~\ref{equ:rep} \cite{chen2020simple} like typical embedders. The model that always uses augmentation is trained under a scenario where augmentation is applied to every query by using Equation~\ref{equ:all} derived by referring to \cite{yan2025o1}. Therefore, the former performs only embedding, while the latter carries out query augmentation before embedding. Details on the training losses are described in Section \ref{subsec:ada}.}. Out of the 20 datasets in MMEB that contain both training and test sets, the model without augmentation showed better or comparable performance to the model that always uses augmentation on the test sets of the 10 datasets. As augmentation is less effective on these 10 datasets, we consider their queries as not requiring augmentation, while regarding the queries of the remaining 10 datasets as requiring augmentation. Consequently, the datasets are divided as follows:
\begin{itemize}
\item{\textit{Requiring Augmentation}}: ChartQA, DocVQA, ImageNet\_1K, InfographicsVQA, MSCOCO, OK-VQA, SUN397, VisDial, Visual7W, HatefulMemes
\item{\textit{Not Requiring Augmentation}}: A-OKVQA, CIRR, MSCOCO\_i2t, MSCOCO\_t2i, N24News, NIGHTS, VisualNews\_i2t, VisualNews\_t2i, VOC2007, WebQA
\end{itemize}

\begin{figure*}[h]
  \centering
  \includegraphics[width=\linewidth]{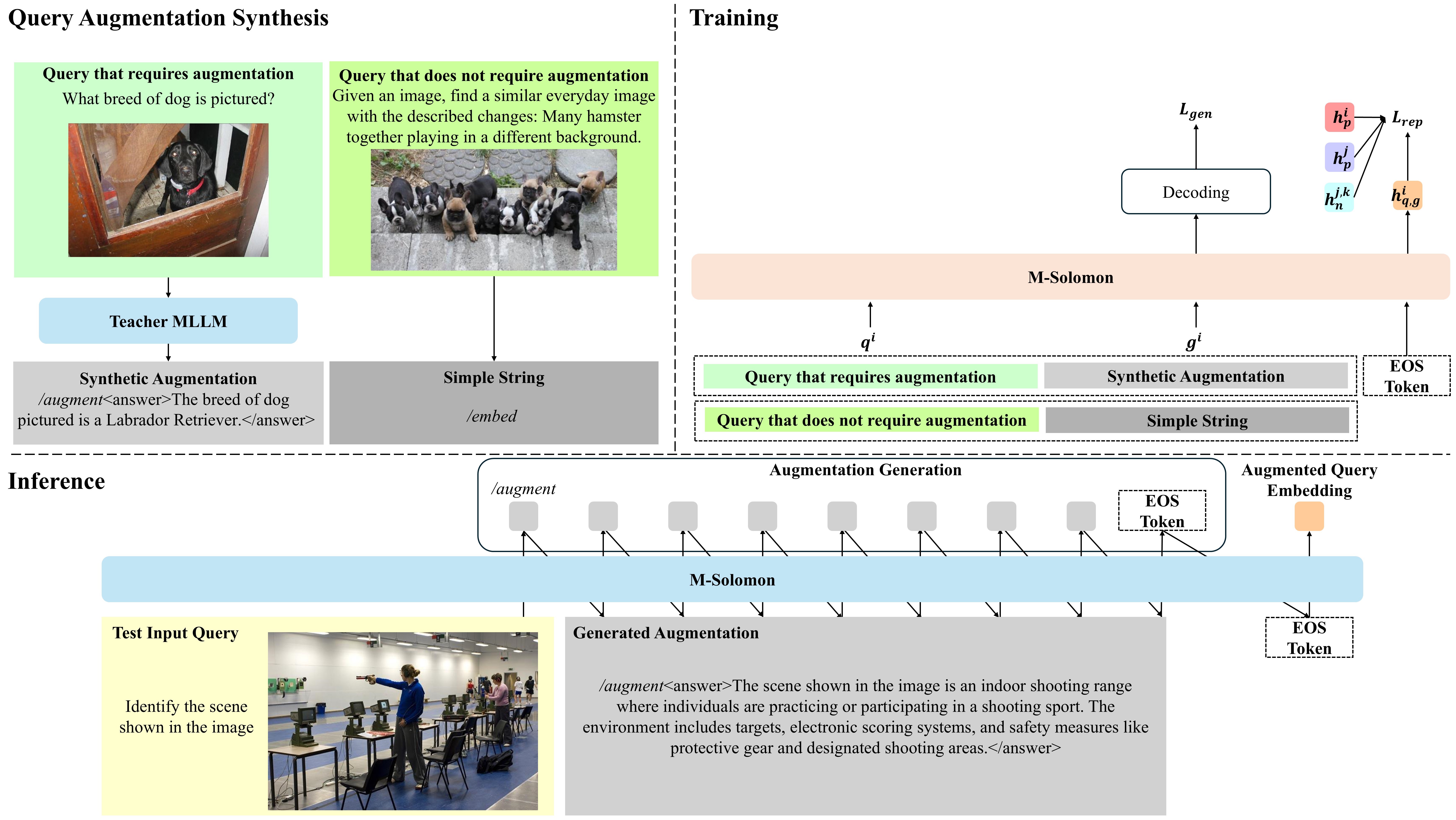}
  \caption{The process of query augmentation synthesis and the training and inference procedure of M-Solomon.}
  \Description{A woman and a girl in white dresses sit in an open car.}
  \label{fig:process}
\end{figure*}

As illustrated in the upper left part of Figure~\ref{fig:process}, for queries that require augmentation, we design a synthesis process to generate augmentations by leveraging Qwen2.5-VL-72B-Instruct, a powerful MLLM as a teacher model \cite{bai2025qwen2}. Following the previous work \cite{yan2025o1}, we regard answers to queries as augmentations because the answers include useful information. In the end, as shown in Table ~\ref{tab:prompt}, we construct a prompt template by referring to the template of \cite{guo2025deepseek} and feed them into the teacher model. The generated outputs include both the reasoning process and the answer, but we only extract and use the answer part.

\begin{table}[t]
    \centering
    \begin{tabular}{|p{0.9\linewidth}|}
        \hline
        The User asks a question (with an image), and the Assistant solves it.\newline
        The assistant first thinks about the reasoning process in the mind and then provides the user with the answer.\newline
        The reasoning process and answer are enclosed within <think> </think> and <answer> </answer> tags, respectively, i.e., <think> reasoning process here </think> <answer> answer here </answer>.\newline
        User: \{question\}. Assistant:
         \\
        \hline
    \end{tabular}
    \caption{Prompt template for query augmentation synthesis. During synthesis, queries will replace \{question\}.}
    \label{tab:prompt}
\end{table}

\subsection{Adaptive Query Augmentation}
\label{subsec:ada}

In the upper right part of Figure~\ref{fig:process}, M-Solomon learns representation for embedding and generation for adaptive query augmentation.

To learn adaptively augmenting queries, M-Solomon is trained to generate synthetic augmentations with the prefix \textit{/augment} for queries that require augmentation, and to generate the simple string \textit{/embed} for those that do not. This allows M-Solomon, given a query, to automatically decide whether to produce \textit{/augment} or \textit{/embed} at the beginning, ultimately enabling M-Solomon to understand when to augment query. If \textit{/augment} is produced, M-Solomon continues to produce augmentation. The objective function for adaptive query augmentation with autoregressive loss is as follows \cite{yu2025cafe}:
\begin{equation}
  \mathcal{L}_{\text{gen}} = - \sum_{t=1}^{T} \log P(g_t \mid q, g_{<t})
  \label{equ:gen}
\end{equation}
where $g$ can be a synthetic augmentation with the prefix \textit{/augment} or the string \textit{/embed}, $t$ is the position for the target token $g_t$, and $P(g_t \mid q, g_{<t})$ is the probability distribution for predicting $g_t$.

To learn representation for embedding, we employ widely used contrastive loss that encourages the anchor embedding to be close to the positive document embedding and distant from the hard negative document embedding \cite{chen2020simple}. The objective function for representation with standard contrastive loss is as follows \cite{jiang2024vlm2vec}:
\begin{equation}
    \mathcal{L}_{\text{rep}} = -\frac{1}{N} \sum_{i=1}^{N} \log \frac{\phi(h^i_{q,g}, h^{i}_{p})}{\sum_{j=1}^{N}(\phi(h^i_{q,g}, h^{j}_{p}) + \sum_{k=1}^{m} \phi(h^i_{q,g},h^{j,k}_{n}))}
  \label{equ:rep}
\end{equation}
where $h_{q,g}, h_p$ and $h_n$ each denote embeddings of the augmented query, the positive document, and the hard negative document. M-Solomon obtain the embeddings from the last hidden states of the eos tokens at the final position. $N$ means the batch size and the function $\phi(\cdot)$ indicates $\exp\left(\cos(\cdot)/\tau\right)$, where $\cos(\cdot)$ and $\tau$ each indicate cosine similarity and temperature hyper-parameter \cite{chen2025mme5}.

The overall objective function is a linear combination of contrastive learning $\mathcal{L}_{\text{rep}}$ and autoregressive learning $\mathcal{L}_{\text{gen}}$ objectives:
\begin{equation}
  \mathcal{L} = \alpha_{\text{rep}} \mathcal{L}_{\text{rep}} + \alpha_{\text{gen}}\mathcal{L}_{\text{gen}}
  \label{equ:all}
\end{equation}
where, $\alpha_{\text{rep}}$ and $\alpha_{\text{gen}}$ are scaling hyper-parameters \cite{yu2025cafe}. We built the overall loss (Equation~\ref{equ:all}) based on \cite{yan2025o1}, however, we did not augment every query. Moreover, contrastive loss on the original queries in addition to the augmented ones was applied in \cite{yan2025o1}, which we found ineffective in our experiments and therefore omitted.

As described in the bottom part of Figure~\ref{fig:process}, during inference, M-Solomon generates \textit{/augment} and augmentations if it determines that queries demand augmentation. Otherwise, it simply generates \textit{/embed}. The generated augmentations are appended to the queries. These augmented queries are then encoded for creating embeddings. Generating and encoding processes are conducted by one-time forward pass. Eventually, the adaptive query augmentation allows M-Solomon to represent the embeddings more informatively and efficiently.

\section{Experiments}
\subsection{Experimental Setup}
\textbf{Datasets and Evaluation Metric.} We employed the training dataset collection of MMEB provided by \cite{chen2025mme5}, which consists of 20 datasets. We used 2.5K queries from each dataset for synthesis process and training, resulting in total 50K training samples. For evaluation, we employed MMEB benchmark that contains 20 in-distribution (IND) and 16 out-of-distribution (OOD) datasets across four task categories - Classification, VQA, Retrieval, and Grounding \cite{jiang2024vlm2vec}. We regarded Precision@1 (P@1) as the primary metric, which is a commonly used metric in MMEB and assigns a score of 1 if the top-ranked retrieved document is relevant, and 0 otherwise. Furthermore, we utilized Latency, \# of $Ts$, and \textit{/embed}\%. Latency measures the average time (ms/query) taken for generation. \# of $Ts$ indicates the average number of generated tokens and $Ts$ denotes `Tokens' \cite{zhang2025adaptthink}. \textit{/embed}\% measures the rate for selecting \textit{/embed}. Also, we used $CF$, which denotes `Confidence' and is calculated by selecting the higher probability between the generations of \textit{/augment} and \textit{/embed} to measure how confidently these tokens are generated.

\textbf{Baselines.} Since our methodology can be easily integrated into various approaches, we primarily compared the models before and after applying adaptive query augmentation to demonstrate its effectiveness. Therefore, as baselines, we used NoAug which does not use augmentation and AlwaysAug which always uses augmentation\footnote{NoAug and AlwaysAug are identical to the model without augmentation and the model that always uses augmentation in Figure~\ref{fig:pilot}, respectively.}. NoAug was trained by using only contrastive loss, while we trained AlwaysAug by setting all training samples to include augmentations and employing Equation~\ref{equ:all}. Lastly, we reported VLM2Vec \cite{jiang2024vlm2vec}, which was based on Qwen2-VL-7B-Instruct \cite{wang2024qwen2} and also trained with contrastive loss. However, unlike our approach, VLM2Vec did not use hard negative documents and utilized 50K queries from each dataset. VLM2Vec provided the reference performance on MMEB benchmark, establishing a lower bound for performance.
For ablation study, we adopt M-Solomon-Half, which was trained by randomly selecting and augmenting half of the queries in each training dataset without dividing datasets in Section~\ref{subsec:synthesis}. Moreover, we presented M-Solomon-\textit{/embed} and M-Solomon-\textit{/augment}, which were obligated to append \textit{/embed} to the query to prevent augmentation and \textit{/augment} to the query to enforce augmentation during inference, respectively.

\textbf{Implementation Details.} M-Solomon was based on Qwen2-VL-7B-Instruct \cite{wang2024qwen2} and was trained and evaluated on a single node with 8×A100 80GB GPUs. We used LoRA with a rank of 16 \cite{hu2022lora}. We set $m, N, \tau, \alpha_{\text{rep}}$, and $\alpha_{\text{gen}}$ as 1, 128, 0.02, 1.0, and 0.1, respectively. Following \cite{jiang2024vlm2vec}, we applied GradCache \cite{gao2021scaling}. The image resolution was fixed at 512×512, and the maximum token length was set to 1800. M-Solomon was trained for 1 epoch with learning rate of 2e-5, linear scheduler, and warmup steps of 0. The baselines were also trained in the same conditions.

\begin{table*}[t]
\centering
\begin{tabular}{lcccc|ccc|ccc}
\toprule
 Models &  Classification & VQA & Retrieval & Grounding & IND & OOD & Overall & Latency & \# of $Ts$ (\textit{/embed}\%) & $CF$ \\
\hline
\# of Datasets & 10 & 10 & 12 & 4 & 20 & 16 & 36 & 36 & 36 & 36 \\
\midrule
VLM2Vec \cite{jiang2024vlm2vec} & 62.6 & 57.8 & \textbf{69.9} & 81.7 & \textbf{72.2} & 57.8 & 65.8 & - & - & - \\
NoAug & 61.9 & 59.6 & 68.1 & 83.9 & 68.7 & 63.1 & 66.1 & - & - & - \\
AlwaysAug & 64.4 & \textbf{62.4} & 67.1 & \textbf{85.5} & 69.9 & 64.7 & 67.4 & 1320 & 45.8 (0\%) & - \\
\hline
M-Solomon (Ours) & 64.4 & 61.9 & 68.8 & 83.6 & 69.6 & \textbf{65.4} & \textbf{67.6} & 716 & 23.8 (55.1\%) & \textbf{93.1} \\
M-Solomon-Half & 62.5 & 62.0 & 68.1 & 84.8 & 69.4 & 64.1 & 67.0 & 771 & 25.9 (51.6\%) & 67.6 \\
M-Solomon-\textit{/embed} & 63.7 & 57.2 & 68.6 & 83.7 & 67.8 & 64.3 & 66.0 & \textbf{93} & 1.0 (100\%) & - \\
M-Solomon-\textit{/augment} & \textbf{64.5} & 62.0 & 67.6 & 83.7 & 69.1 & 65.4 & 67.3 & 655 & 20.9 (0\%) & - \\
\bottomrule
\end{tabular}
\caption{Results on MMEB benchmark. The scores are averaged based on the conditions defined for each column. If the metric is not specified, the default is P@1.}
\label{tab:results}
\end{table*}

\subsection{Main Results}
The performance results of the models are presented in Table~\ref{tab:results}. First, NoAug, AlwaysAug, and M-Solomon all achieved higher overall performance than VLM2Vec, which used the 662K samples without hard negative documents. This showed that effective embedders could be developed with a smaller number of samples and highlighted the importance of using hard negative documents in embedding tasks. Both AlwaysAug and M-Solomon surpassed NoAug on overall performance by a large margin, indicating that answer-style augmentation provided useful information. However, AlwaysAug exhibited lower performance than M-Solomon. Even Latency and \# of $Ts$ were nearly twice as high, showing that embedding latency of AlwaysAug was significantly slower. This was because M-Solomon adaptively generated augmentations when it determined that queries required them. Moreover, since \textit{/embed}\% of M-Solomon was close to 50\%, we could observe that \textit{/augment} and \textit{/embed} were selected evenly, indicating that M-Solomon performed adaptive query augmentation appropriately. This adaptive and balanced selection between \textit{/augment} and \textit{/embed} was not random but deliberately made by M-Solomon because $CF$ of M-Solomon was substantially high. Lastly, M-Solomon notably outperformed NoAug and AlwaysAug on OOD result, which demonstrated the generalization effect of adaptive query augmentation.

\subsection{Ablation Study}
As shown in the bottom part of Table~\ref{tab:results}, M-Solomon-Half resulted in lower overall performance and higher Latency and \# of $Ts$ than M-Solomon. This highlighted the importance of identifying training datasets that require augmentation. Also, M-Solomon-Half produced less confident augmentations with reduced $CF$ score. Despite the enforcement of \textit{/augment}, M-Solomon-\textit{/augment} unexpectedly exhibited lower Latency and \# of $Ts$ than M-Solomon because it abnormally halted generation due to conflicts caused by appending \textit{/augment} to queries that did not require augmentation. Ultimately, M-Solomon-\textit{/augment} showed lower overall performance than M-Solomon, indicating that the suppression of adaptive query augmentation was not beneficial. M-Solomon-\textit{/embed} created embeddings quickly in the absence of augmentation, which happened due to the enforcement of \textit{/embed} and the conflicts caused by appending \textit{/embed} to queries that required augmentation. However, the overall effectiveness was inferior.

\begin{table}[t]
\centering
\begin{tabular}{lcccc}
\toprule
Models & P@1 & Latency & \# of $Ts$ (\textit{/embed}\%) & $CF$ \\
\midrule
\textbf{FashionIQ} \\
\hline
NoAug & 23.1 & - & - & - \\
AlwaysAug & 21.1 & 1496 & 51.6 (0\%) & - \\
M-Solomon & \textbf{26.7} & \textbf{333} & 9.2 (91.0\%) & \textbf{80.6} \\
\hline
\textbf{GQA} \\
\hline
NoAug & 61.5 & - & - & - \\
AlwaysAug & 64.3 & \textbf{497} & 15.4 (0\%) & - \\
M-Solomon & \textbf{68.1} & 663 & 21.3 (8.3\%) & \textbf{88.8} \\
\hline
\textbf{ImageNet-R} \\
\hline
NoAug & 85.3 & - & - & - \\
AlwaysAug & 88.5 & 1292 & 43.8 (0\%) & - \\
M-Solomon & \textbf{90.3} & \textbf{1266} & 43.4 (1.0\%) & \textbf{97.0} \\
\bottomrule
\end{tabular}
\caption{Results on the several datasets. The scores are averaged for each dataset.}
\label{tab:further}
\end{table}

\subsection{Further Analysis of Adaptive Query Augmentation}
In this subsection, as presented in Table~\ref{tab:further}, we further analyzed how M-Solomon generated augmentations more adaptively than NoAug and AlwaysAug across FashionIQ, GQA, and ImageNet-R OOD datasets in MMEB benchmark. While M-Solomon consistently demonstrated strong performance on other datasets, we presented the results on three representative datasets. To begin with, on all three datasets, M-Solomon significantly outperformed NoAug. However, AlwaysAug performed worse than NoAug on FashionIQ, and on other datasets its improvement over NoAug was smaller compared to M-Solomon. This demonstrated the effectiveness of adaptive query augmentation.

In FashionIQ dataset, M-Solomon significantly outperformed AlwaysAug on P@1 score, achieving approximately five times lower Latency and \# of $Ts$. As \textit{/embed}\% was 91.0\%, M-Solomon determined that most queries of FashionIQ dataset did not require augmentation, which contributed to superior results and embedding latency of M-Solomon.

In GQA dataset, M-Solomon achieved higher performance than AlwaysAug on P@1. However, despite \textit{/embed}\% of 8.3\% that could reduce Latency, M-Solomon showed slightly higher Latency and \# of $Ts$. This was because M-Solomon generated longer and more informative augmentations that led to stronger performance. As shown in the first GQA example of Table~\ref{tab:example}, while AlwaysAug merely generated `No' for the given query, which caused AlwaysAug to find a irrelevant document, M-Solomon produced a longer and helpful augmentation, specifically highlighting that the boat is on the left in the picture, which enabled M-Solomon to retrieve a correct document. It was expected that this capability arose from learning to appropriately generate augmentations only when they were required by queries, rather than generating them for all queries.

In ImageNet-R dataset, AlwaysAug and M-Solomon revealed similar scores on Latency and \# of $Ts$, while M-Solomon exhibited notably higher P@1 score. This further demonstrated that M-Solomon performed more meaningful query augmentation. In the second ImageNet-R example of Table~\ref{tab:example}, while AlwaysAug performed augmentation under the misconception that the given query image depicted a hedgehog or a spiny animal, M-Solomon generated augmentation by recognizing the image as a fluffy, dog-like animal, which helped to retrieve a relevant document.

\begin{table}[t]
\centering
\begin{tabular}{ll}
\toprule
\textbf{GQA} \\
\hline
Query & Is the boat on the right of the picture? \\
& \includegraphics[width=5cm]{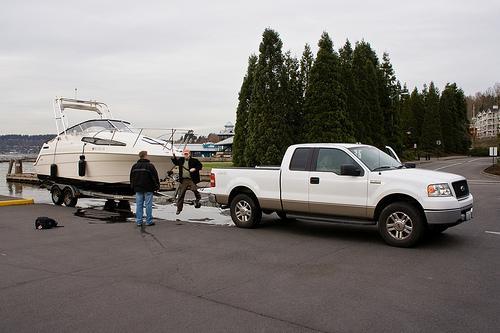}\\
\hline
\textit{aug} of AlwaysAug & No \\
Top 1 Document & No, there are no boats. \\
\hline
\textit{aug} of M-Solomon & No, the boat is on the left of the picture. \\
Top 1 Document & No, the boat is on the left of the image. \\
\hline
Relevant Document & No, the boat is on the left of the image. \\
\hline
\textbf{ImageNet-R} \\
\hline
Query & Represent the given image for \\
& classification \\
& \includegraphics[width=2.5cm]{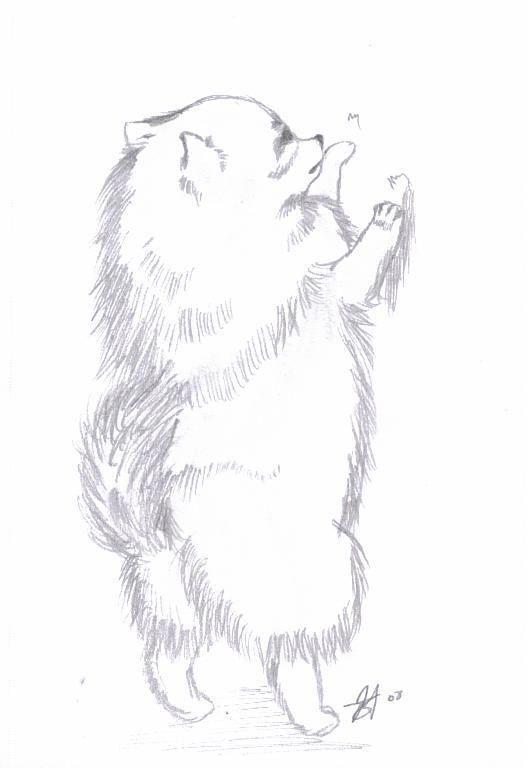}\\
\hline
\textit{aug} of AlwaysAug & The image depicts a cartoonish \\
& anthropomorphic creature resembling \\
& a hedgehog or a small animal with \\
& spines-like features, standing upright \\
& on its hind legs and holding \\ 
& its front paws up as if it is dancing \\
& or performing a trick. The creature \\
& has a fluffy appearance and is \\
& drawn in a simplistic sketch-like style. \\
Top 1 Document & porcupine \\
\hline
\textit{aug} of M-Solomon & The image depicts a stylized, \\ 
& cartoon-like drawing of a fluffy animal, \\
& likely a dog, standing on its hind legs \\
& with its front paws raised. The drawing \\
& is monochromatic, using shades of gray, \\
& and appears to be a sketch or a \\
& preliminary drawing. \\
Top 1 Document & pomeranian \\
\hline
Relevant Document & pomeranian \\
\hline
\end{tabular}
\caption{Augmentation and retrieval examples of AlwaysAug and M-Solomon. \textit{aug} means augmentation. In the examples, M-Solomon generated higher-quality augmentations, leading to more accurate retrieval results.}
\label{tab:example}
\end{table}

Across all datasets, M-Solomon consistently achieved robust $CF$ scores, which showed that it conducted adaptive query augmentation confidently and accurately.

\section{Conclusion and Future Work}
In this work, we proposed M-Solomon, a universal multimodal embedder that can adaptively determine when to augment queries. We first identified queries that require augmentation at the dataset level and synthesized augmentations for those queries. Then, M-Solomon was trained to generate synthetic augmentations with the prefix \textit{/augment} or the simple string \textit{/embed} based on whether queries demanded augmentation, which enabled M-Solomon to understand when to augment queries. Experimental results showed that M-Solomon significantly outperformed the baselines with effectively and efficiently creating embeddings, which demonstrated the validity of our approach. In the future, we will study methods to identify which queries require augmentation at the query level because this allows for precise decisions by reflecting fine-grained information of each query. Furthermore, we will extend adaptive query augmentation with another option that performs reasoning-based query augmentation for reasoning-intensive embedding tasks such as BRIGHT \cite{su2024bright} and RAR-b \cite{xiao2024rar}.

\section{GenAI Usage Disclosure}
\subsection{Usage in Research Stage}
We leveraged Qwen2.5-VL-72B-Instruct\footnote{\url{https://huggingface.co/Qwen/Qwen2.5-VL-72B-Instruct}} \cite{bai2025qwen2}, a powerful and publicly available MLLM as a teacher model for query augmentation synthesis. By utilizing the template of \cite{guo2025deepseek}, we constructed prompts and fed them into the teacher model to obtain answer-style augmentations. The teacher model generated corresponding augmentations for total 50K queries.
\subsection{Usage in Writing Stage}
During writing, we occasionally used Generative AI like ChatGPT\footnote{\url{https://chatgpt.com/}} for basic and straightforward tasks such as translation, finding synonyms, refining grammar, checking spell, and correcting awkward or incorrect expressions. Even though we obtained outputs from Generative AI for such purposes, we carefully checked and revised them before using their use. Moreover, all the content of this paper was initially written and created on our own without Generative AI. We believe that the use of Generative AI to this extent is acceptable and can strongly support active research and paper writing in a positive way.

\begin{acks}
This work was supported by Institute for Information \& communications Technology Promotion(IITP) grant funded by the Korea government(MSIT) (RS-2024-00398115, Research on the reliability and coherence of outcomes produced by Generative AI).
\end{acks}

\bibliographystyle{ACM-Reference-Format}
\bibliography{sample-base}


\begin{thebibliography}{30}


\ifx \showCODEN    \undefined \def \showCODEN     #1{\unskip}     \fi
\ifx \showISBNx    \undefined \def \showISBNx     #1{\unskip}     \fi
\ifx \showISBNxiii \undefined \def \showISBNxiii  #1{\unskip}     \fi
\ifx \showISSN     \undefined \def \showISSN      #1{\unskip}     \fi
\ifx \showLCCN     \undefined \def \showLCCN      #1{\unskip}     \fi
\ifx \shownote     \undefined \def \shownote      #1{#1}          \fi
\ifx \showarticletitle \undefined \def \showarticletitle #1{#1}   \fi
\ifx \showURL      \undefined \def \showURL       {\relax}        \fi
\providecommand\bibfield[2]{#2}
\providecommand\bibinfo[2]{#2}
\providecommand\natexlab[1]{#1}
\providecommand\showeprint[2][]{arXiv:#2}

\bibitem[Bai et~al\mbox{.}(2025)]%
        {bai2025qwen2}
\bibfield{author}{\bibinfo{person}{Shuai Bai}, \bibinfo{person}{Keqin Chen}, \bibinfo{person}{Xuejing Liu}, \bibinfo{person}{Jialin Wang}, \bibinfo{person}{Wenbin Ge}, \bibinfo{person}{Sibo Song}, \bibinfo{person}{Kai Dang}, \bibinfo{person}{Peng Wang}, \bibinfo{person}{Shijie Wang}, \bibinfo{person}{Jun Tang}, {et~al\mbox{.}}} \bibinfo{year}{2025}\natexlab{}.
\newblock \showarticletitle{Qwen2. 5-vl technical report}.
\newblock \bibinfo{journal}{\emph{arXiv preprint arXiv:2502.13923}} (\bibinfo{year}{2025}).
\newblock


\bibitem[Chen et~al\mbox{.}(2025)]%
        {chen2025mme5}
\bibfield{author}{\bibinfo{person}{Haonan Chen}, \bibinfo{person}{Liang Wang}, \bibinfo{person}{Nan Yang}, \bibinfo{person}{Yutao Zhu}, \bibinfo{person}{Ziliang Zhao}, \bibinfo{person}{Furu Wei}, {and} \bibinfo{person}{Zhicheng Dou}.} \bibinfo{year}{2025}\natexlab{}.
\newblock \showarticletitle{mmE5: Improving Multimodal Multilingual Embeddings via High-quality Synthetic Data}.
\newblock \bibinfo{journal}{\emph{arXiv preprint arXiv:2502.08468}} (\bibinfo{year}{2025}).
\newblock


\bibitem[Chen et~al\mbox{.}(2020)]%
        {chen2020simple}
\bibfield{author}{\bibinfo{person}{Ting Chen}, \bibinfo{person}{Simon Kornblith}, \bibinfo{person}{Mohammad Norouzi}, {and} \bibinfo{person}{Geoffrey Hinton}.} \bibinfo{year}{2020}\natexlab{}.
\newblock \showarticletitle{A simple framework for contrastive learning of visual representations}. In \bibinfo{booktitle}{\emph{International conference on machine learning}}. PmLR, \bibinfo{pages}{1597--1607}.
\newblock


\bibitem[Fang et~al\mbox{.}(2025)]%
        {fang2025thinkless}
\bibfield{author}{\bibinfo{person}{Gongfan Fang}, \bibinfo{person}{Xinyin Ma}, {and} \bibinfo{person}{Xinchao Wang}.} \bibinfo{year}{2025}\natexlab{}.
\newblock \showarticletitle{Thinkless: LLM Learns When to Think}.
\newblock \bibinfo{journal}{\emph{arXiv preprint arXiv:2505.13379}} (\bibinfo{year}{2025}).
\newblock


\bibitem[Gao et~al\mbox{.}(2021)]%
        {gao2021scaling}
\bibfield{author}{\bibinfo{person}{Luyu Gao}, \bibinfo{person}{Yunyi Zhang}, \bibinfo{person}{Jiawei Han}, {and} \bibinfo{person}{Jamie Callan}.} \bibinfo{year}{2021}\natexlab{}.
\newblock \showarticletitle{Scaling deep contrastive learning batch size under memory limited setup}.
\newblock \bibinfo{journal}{\emph{arXiv preprint arXiv:2101.06983}} (\bibinfo{year}{2021}).
\newblock


\bibitem[Gu et~al\mbox{.}(2025)]%
        {gu2025breaking}
\bibfield{author}{\bibinfo{person}{Tiancheng Gu}, \bibinfo{person}{Kaicheng Yang}, \bibinfo{person}{Ziyong Feng}, \bibinfo{person}{Xingjun Wang}, \bibinfo{person}{Yanzhao Zhang}, \bibinfo{person}{Dingkun Long}, \bibinfo{person}{Yingda Chen}, \bibinfo{person}{Weidong Cai}, {and} \bibinfo{person}{Jiankang Deng}.} \bibinfo{year}{2025}\natexlab{}.
\newblock \showarticletitle{Breaking the Modality Barrier: Universal Embedding Learning with Multimodal LLMs}.
\newblock \bibinfo{journal}{\emph{arXiv preprint arXiv:2504.17432}} (\bibinfo{year}{2025}).
\newblock


\bibitem[Guo et~al\mbox{.}(2025)]%
        {guo2025deepseek}
\bibfield{author}{\bibinfo{person}{Daya Guo}, \bibinfo{person}{Dejian Yang}, \bibinfo{person}{Haowei Zhang}, \bibinfo{person}{Junxiao Song}, \bibinfo{person}{Ruoyu Zhang}, \bibinfo{person}{Runxin Xu}, \bibinfo{person}{Qihao Zhu}, \bibinfo{person}{Shirong Ma}, \bibinfo{person}{Peiyi Wang}, \bibinfo{person}{Xiao Bi}, {et~al\mbox{.}}} \bibinfo{year}{2025}\natexlab{}.
\newblock \showarticletitle{Deepseek-r1: Incentivizing reasoning capability in llms via reinforcement learning}.
\newblock \bibinfo{journal}{\emph{arXiv preprint arXiv:2501.12948}} (\bibinfo{year}{2025}).
\newblock


\bibitem[Jiang et~al\mbox{.}(2025)]%
        {jiang2025think}
\bibfield{author}{\bibinfo{person}{Lingjie Jiang}, \bibinfo{person}{Xun Wu}, \bibinfo{person}{Shaohan Huang}, \bibinfo{person}{Qingxiu Dong}, \bibinfo{person}{Zewen Chi}, \bibinfo{person}{Li Dong}, \bibinfo{person}{Xingxing Zhang}, \bibinfo{person}{Tengchao Lv}, \bibinfo{person}{Lei Cui}, {and} \bibinfo{person}{Furu Wei}.} \bibinfo{year}{2025}\natexlab{}.
\newblock \showarticletitle{Think Only When You Need with Large Hybrid-Reasoning Models}.
\newblock \bibinfo{journal}{\emph{arXiv preprint arXiv:2505.14631}} (\bibinfo{year}{2025}).
\newblock


\bibitem[Jiang et~al\mbox{.}(2024)]%
        {jiang2024vlm2vec}
\bibfield{author}{\bibinfo{person}{Ziyan Jiang}, \bibinfo{person}{Rui Meng}, \bibinfo{person}{Xinyi Yang}, \bibinfo{person}{Semih Yavuz}, \bibinfo{person}{Yingbo Zhou}, {and} \bibinfo{person}{Wenhu Chen}.} \bibinfo{year}{2024}\natexlab{}.
\newblock \showarticletitle{Vlm2vec: Training vision-language models for massive multimodal embedding tasks}.
\newblock \bibinfo{journal}{\emph{arXiv preprint arXiv:2410.05160}} (\bibinfo{year}{2024}).
\newblock


\bibitem[Kong et~al\mbox{.}(2025)]%
        {kong2025modality}
\bibfield{author}{\bibinfo{person}{Fanheng Kong}, \bibinfo{person}{Jingyuan Zhang}, \bibinfo{person}{Yahui Liu}, \bibinfo{person}{Hongzhi Zhang}, \bibinfo{person}{Shi Feng}, \bibinfo{person}{Xiaocui Yang}, \bibinfo{person}{Daling Wang}, \bibinfo{person}{Yu Tian}, \bibinfo{person}{Qi Wang}, \bibinfo{person}{Fuzheng Zhang}, {et~al\mbox{.}}} \bibinfo{year}{2025}\natexlab{}.
\newblock \showarticletitle{Modality Curation: Building Universal Embeddings for Advanced Multimodal Information Retrieval}.
\newblock \bibinfo{journal}{\emph{arXiv preprint arXiv:2505.19650}} (\bibinfo{year}{2025}).
\newblock


\bibitem[Lan et~al\mbox{.}(2025)]%
        {lan2025llave}
\bibfield{author}{\bibinfo{person}{Zhibin Lan}, \bibinfo{person}{Liqiang Niu}, \bibinfo{person}{Fandong Meng}, \bibinfo{person}{Jie Zhou}, {and} \bibinfo{person}{Jinsong Su}.} \bibinfo{year}{2025}\natexlab{}.
\newblock \showarticletitle{LLaVE: Large Language and Vision Embedding Models with Hardness-Weighted Contrastive Learning}.
\newblock \bibinfo{journal}{\emph{arXiv preprint arXiv:2503.04812}} (\bibinfo{year}{2025}).
\newblock


\bibitem[Liu et~al\mbox{.}(2025)]%
        {liu2025idmr}
\bibfield{author}{\bibinfo{person}{Bangwei Liu}, \bibinfo{person}{Yicheng Bao}, \bibinfo{person}{Shaohui Lin}, \bibinfo{person}{Xuhong Wang}, \bibinfo{person}{Xin Tan}, \bibinfo{person}{Yingchun Wang}, \bibinfo{person}{Yuan Xie}, {and} \bibinfo{person}{Chaochao Lu}.} \bibinfo{year}{2025}\natexlab{}.
\newblock \showarticletitle{Idmr: Towards instance-driven precise visual correspondence in multimodal retrieval}.
\newblock \bibinfo{journal}{\emph{arXiv preprint arXiv:2504.00954}} (\bibinfo{year}{2025}).
\newblock


\bibitem[Lou et~al\mbox{.}(2025)]%
        {lou2025adacot}
\bibfield{author}{\bibinfo{person}{Chenwei Lou}, \bibinfo{person}{Zewei Sun}, \bibinfo{person}{Xinnian Liang}, \bibinfo{person}{Meng Qu}, \bibinfo{person}{Wei Shen}, \bibinfo{person}{Wenqi Wang}, \bibinfo{person}{Yuntao Li}, \bibinfo{person}{Qingping Yang}, {and} \bibinfo{person}{Shuangzhi Wu}.} \bibinfo{year}{2025}\natexlab{}.
\newblock \showarticletitle{AdaCoT: Pareto-Optimal Adaptive Chain-of-Thought Triggering via Reinforcement Learning}.
\newblock \bibinfo{journal}{\emph{arXiv preprint arXiv:2505.11896}} (\bibinfo{year}{2025}).
\newblock


\bibitem[Ma et~al\mbox{.}(2025)]%
        {ma2025reasoning}
\bibfield{author}{\bibinfo{person}{Wenjie Ma}, \bibinfo{person}{Jingxuan He}, \bibinfo{person}{Charlie Snell}, \bibinfo{person}{Tyler Griggs}, \bibinfo{person}{Sewon Min}, {and} \bibinfo{person}{Matei Zaharia}.} \bibinfo{year}{2025}\natexlab{}.
\newblock \showarticletitle{Reasoning Models Can Be Effective Without Thinking}.
\newblock \bibinfo{journal}{\emph{arXiv preprint arXiv:2504.09858}} (\bibinfo{year}{2025}).
\newblock


\bibitem[Qin et~al\mbox{.}(2025)]%
        {qin2025unimoco}
\bibfield{author}{\bibinfo{person}{Jiajun Qin}, \bibinfo{person}{Yuan Pu}, \bibinfo{person}{Zhuolun He}, \bibinfo{person}{Seunggeun Kim}, \bibinfo{person}{David~Z Pan}, {and} \bibinfo{person}{Bei Yu}.} \bibinfo{year}{2025}\natexlab{}.
\newblock \showarticletitle{UniMoCo: Unified Modality Completion for Robust Multi-Modal Embeddings}.
\newblock \bibinfo{journal}{\emph{arXiv preprint arXiv:2505.11815}} (\bibinfo{year}{2025}).
\newblock


\bibitem[Schneider et~al\mbox{.}(2025)]%
        {schneider2025abc}
\bibfield{author}{\bibinfo{person}{Benjamin Schneider}, \bibinfo{person}{Florian Kerschbaum}, {and} \bibinfo{person}{Wenhu Chen}.} \bibinfo{year}{2025}\natexlab{}.
\newblock \showarticletitle{ABC: Achieving Better Control of Multimodal Embeddings using VLMs}.
\newblock \bibinfo{journal}{\emph{arXiv preprint arXiv:2503.00329}} (\bibinfo{year}{2025}).
\newblock


\bibitem[Shen et~al\mbox{.}({[n.\,d.]})]%
        {hu2022lora}
\bibfield{author}{\bibinfo{person}{Yelong Shen}, \bibinfo{person}{Phillip Wallis}, \bibinfo{person}{Zeyuan Allen-Zhu}, \bibinfo{person}{Yuanzhi Li}, \bibinfo{person}{Shean Wang}, {et~al\mbox{.}}} \bibinfo{year}{[n.\,d.]}\natexlab{}.
\newblock \showarticletitle{Lora: Low-rank adaptation of large language models.}
\newblock  (\bibinfo{year}{[n.\,d.]}).
\newblock


\bibitem[Su et~al\mbox{.}(2024)]%
        {su2024bright}
\bibfield{author}{\bibinfo{person}{Hongjin Su}, \bibinfo{person}{Howard Yen}, \bibinfo{person}{Mengzhou Xia}, \bibinfo{person}{Weijia Shi}, \bibinfo{person}{Niklas Muennighoff}, \bibinfo{person}{Han-yu Wang}, \bibinfo{person}{Haisu Liu}, \bibinfo{person}{Quan Shi}, \bibinfo{person}{Zachary~S Siegel}, \bibinfo{person}{Michael Tang}, {et~al\mbox{.}}} \bibinfo{year}{2024}\natexlab{}.
\newblock \showarticletitle{Bright: A realistic and challenging benchmark for reasoning-intensive retrieval}.
\newblock \bibinfo{journal}{\emph{arXiv preprint arXiv:2407.12883}} (\bibinfo{year}{2024}).
\newblock


\bibitem[Tang et~al\mbox{.}(2025)]%
        {tang2025think}
\bibfield{author}{\bibinfo{person}{Jiakai Tang}, \bibinfo{person}{Sunhao Dai}, \bibinfo{person}{Teng Shi}, \bibinfo{person}{Jun Xu}, \bibinfo{person}{Xu Chen}, \bibinfo{person}{Wen Chen}, \bibinfo{person}{Wu Jian}, {and} \bibinfo{person}{Yuning Jiang}.} \bibinfo{year}{2025}\natexlab{}.
\newblock \showarticletitle{Think before recommend: Unleashing the latent reasoning power for sequential recommendation}.
\newblock \bibinfo{journal}{\emph{arXiv preprint arXiv:2503.22675}} (\bibinfo{year}{2025}).
\newblock


\bibitem[Thirukovalluru et~al\mbox{.}(2025)]%
        {thirukovalluru2025breaking}
\bibfield{author}{\bibinfo{person}{Raghuveer Thirukovalluru}, \bibinfo{person}{Rui Meng}, \bibinfo{person}{Ye Liu}, \bibinfo{person}{Mingyi Su}, \bibinfo{person}{Ping Nie}, \bibinfo{person}{Semih Yavuz}, \bibinfo{person}{Yingbo Zhou}, \bibinfo{person}{Wenhu Chen}, \bibinfo{person}{Bhuwan Dhingra}, {et~al\mbox{.}}} \bibinfo{year}{2025}\natexlab{}.
\newblock \showarticletitle{Breaking the Batch Barrier (B3) of Contrastive Learning via Smart Batch Mining}.
\newblock \bibinfo{journal}{\emph{arXiv preprint arXiv:2505.11293}} (\bibinfo{year}{2025}).
\newblock


\bibitem[Wang et~al\mbox{.}(2024)]%
        {wang2024qwen2}
\bibfield{author}{\bibinfo{person}{Peng Wang}, \bibinfo{person}{Shuai Bai}, \bibinfo{person}{Sinan Tan}, \bibinfo{person}{Shijie Wang}, \bibinfo{person}{Zhihao Fan}, \bibinfo{person}{Jinze Bai}, \bibinfo{person}{Keqin Chen}, \bibinfo{person}{Xuejing Liu}, \bibinfo{person}{Jialin Wang}, \bibinfo{person}{Wenbin Ge}, {et~al\mbox{.}}} \bibinfo{year}{2024}\natexlab{}.
\newblock \showarticletitle{Qwen2-vl: Enhancing vision-language model's perception of the world at any resolution}.
\newblock \bibinfo{journal}{\emph{arXiv preprint arXiv:2409.12191}} (\bibinfo{year}{2024}).
\newblock


\bibitem[Wang et~al\mbox{.}(2025)]%
        {wang2025pats}
\bibfield{author}{\bibinfo{person}{Yi Wang}, \bibinfo{person}{Junxiao Liu}, \bibinfo{person}{Shimao Zhang}, \bibinfo{person}{Jiajun Chen}, {and} \bibinfo{person}{Shujian Huang}.} \bibinfo{year}{2025}\natexlab{}.
\newblock \showarticletitle{PATS: Process-Level Adaptive Thinking Mode Switching}.
\newblock \bibinfo{journal}{\emph{arXiv preprint arXiv:2505.19250}} (\bibinfo{year}{2025}).
\newblock


\bibitem[Xiao et~al\mbox{.}(2024)]%
        {xiao2024rar}
\bibfield{author}{\bibinfo{person}{Chenghao Xiao}, \bibinfo{person}{G~Thomas Hudson}, {and} \bibinfo{person}{Noura~Al Moubayed}.} \bibinfo{year}{2024}\natexlab{}.
\newblock \showarticletitle{Rar-b: Reasoning as retrieval benchmark}.
\newblock \bibinfo{journal}{\emph{arXiv preprint arXiv:2404.06347}} (\bibinfo{year}{2024}).
\newblock


\bibitem[Xue et~al\mbox{.}(2025)]%
        {xue2025improve}
\bibfield{author}{\bibinfo{person}{Youze Xue}, \bibinfo{person}{Dian Li}, {and} \bibinfo{person}{Gang Liu}.} \bibinfo{year}{2025}\natexlab{}.
\newblock \showarticletitle{Improve Multi-Modal Embedding Learning via Explicit Hard Negative Gradient Amplifying}.
\newblock \bibinfo{journal}{\emph{arXiv preprint arXiv:2506.02020}} (\bibinfo{year}{2025}).
\newblock


\bibitem[Yan et~al\mbox{.}(2025)]%
        {yan2025o1}
\bibfield{author}{\bibinfo{person}{Ruiran Yan}, \bibinfo{person}{Zheng Liu}, {and} \bibinfo{person}{Defu Lian}.} \bibinfo{year}{2025}\natexlab{}.
\newblock \showarticletitle{O1 embedder: Let retrievers think before action}.
\newblock \bibinfo{journal}{\emph{arXiv preprint arXiv:2502.07555}} (\bibinfo{year}{2025}).
\newblock


\bibitem[Yang et~al\mbox{.}(2025)]%
        {yang2025qwen3}
\bibfield{author}{\bibinfo{person}{An Yang}, \bibinfo{person}{Anfeng Li}, \bibinfo{person}{Baosong Yang}, \bibinfo{person}{Beichen Zhang}, \bibinfo{person}{Binyuan Hui}, \bibinfo{person}{Bo Zheng}, \bibinfo{person}{Bowen Yu}, \bibinfo{person}{Chang Gao}, \bibinfo{person}{Chengen Huang}, \bibinfo{person}{Chenxu Lv}, {et~al\mbox{.}}} \bibinfo{year}{2025}\natexlab{}.
\newblock \showarticletitle{Qwen3 technical report}.
\newblock \bibinfo{journal}{\emph{arXiv preprint arXiv:2505.09388}} (\bibinfo{year}{2025}).
\newblock


\bibitem[Yu et~al\mbox{.}(2025)]%
        {yu2025cafe}
\bibfield{author}{\bibinfo{person}{Hao Yu}, \bibinfo{person}{Zhuokai Zhao}, \bibinfo{person}{Shen Yan}, \bibinfo{person}{Lukasz Korycki}, \bibinfo{person}{Jianyu Wang}, \bibinfo{person}{Baosheng He}, \bibinfo{person}{Jiayi Liu}, \bibinfo{person}{Lizhu Zhang}, \bibinfo{person}{Xiangjun Fan}, {and} \bibinfo{person}{Hanchao Yu}.} \bibinfo{year}{2025}\natexlab{}.
\newblock \showarticletitle{CAFe: Unifying Representation and Generation with Contrastive-Autoregressive Finetuning}.
\newblock \bibinfo{journal}{\emph{arXiv preprint arXiv:2503.19900}} (\bibinfo{year}{2025}).
\newblock


\bibitem[Yue et~al\mbox{.}(2025)]%
        {yue2025hybrid}
\bibfield{author}{\bibinfo{person}{Zhenrui Yue}, \bibinfo{person}{Bowen Jin}, \bibinfo{person}{Huimin Zeng}, \bibinfo{person}{Honglei Zhuang}, \bibinfo{person}{Zhen Qin}, \bibinfo{person}{Jinsung Yoon}, \bibinfo{person}{Lanyu Shang}, \bibinfo{person}{Jiawei Han}, {and} \bibinfo{person}{Dong Wang}.} \bibinfo{year}{2025}\natexlab{}.
\newblock \showarticletitle{Hybrid Latent Reasoning via Reinforcement Learning}.
\newblock \bibinfo{journal}{\emph{arXiv preprint arXiv:2505.18454}} (\bibinfo{year}{2025}).
\newblock


\bibitem[Zhang et~al\mbox{.}(2025)]%
        {zhang2025adaptthink}
\bibfield{author}{\bibinfo{person}{Jiajie Zhang}, \bibinfo{person}{Nianyi Lin}, \bibinfo{person}{Lei Hou}, \bibinfo{person}{Ling Feng}, {and} \bibinfo{person}{Juanzi Li}.} \bibinfo{year}{2025}\natexlab{}.
\newblock \showarticletitle{Adaptthink: Reasoning models can learn when to think}.
\newblock \bibinfo{journal}{\emph{arXiv preprint arXiv:2505.13417}} (\bibinfo{year}{2025}).
\newblock


\bibitem[Zhou et~al\mbox{.}(2024)]%
        {zhou2024megapairs}
\bibfield{author}{\bibinfo{person}{Junjie Zhou}, \bibinfo{person}{Zheng Liu}, \bibinfo{person}{Ze Liu}, \bibinfo{person}{Shitao Xiao}, \bibinfo{person}{Yueze Wang}, \bibinfo{person}{Bo Zhao}, \bibinfo{person}{Chen~Jason Zhang}, \bibinfo{person}{Defu Lian}, {and} \bibinfo{person}{Yongping Xiong}.} \bibinfo{year}{2024}\natexlab{}.
\newblock \showarticletitle{MegaPairs: Massive Data Synthesis For Universal Multimodal Retrieval}.
\newblock \bibinfo{journal}{\emph{arXiv preprint arXiv:2412.14475}} (\bibinfo{year}{2024}).
\newblock


\end{thebibliography}










\end{document}